\documentclass[runningheads]{llncs}
\usepackage{times}
\usepackage{epsfig}
\usepackage{graphicx}
\usepackage{amsmath}
\usepackage{amssymb}
\usepackage{times}
\usepackage{soul}
\usepackage{url}
\usepackage[hidelinks]{hyperref}
\usepackage[utf8]{inputenc}
\usepackage[small]{caption}
\usepackage{graphicx}
\usepackage{amsmath}
\usepackage{booktabs}
\usepackage{multirow}
\usepackage{amsfonts}
\usepackage{multirow}
\usepackage{color}
\usepackage[table]{xcolor}
\definecolor{lightgray}{gray}{0.9}
\definecolor{grey1}{gray}{0.8}
\usepackage{wrapfig,lipsum,booktabs}
\usepackage[symbol]{footmisc}

\DeclareMathOperator*{\argmin}{arg\,min}
\title{SeizureNet: Multi-Spectral Deep Feature Learning for Seizure Type Classification}
\author{Umar Asif\inst{1} \and
	Subhrajit Roy\inst{1\thanks{now with Google}} \and
	Jianbin Tang\inst{1} \and
	Stefan Harrer\inst{1}}
\authorrunning{U. Asif et al.}
\institute{IBM Research Australia\\
	\email{umarasif@au1.ibm.com}}
\begin{document}
\maketitle
\begin{abstract}
Automatic classification of epileptic seizure types in electroencephalograms (EEGs) data can enable more precise diagnosis and efficient management of the disease.
This task is challenging due to factors such as low signal-to-noise ratios, signal artefacts, high variance in seizure semiology among epileptic patients, and limited availability of clinical data. 
To overcome these challenges, in this paper, we present SeizureNet, a deep learning framework which learns multi-spectral feature embeddings using an ensemble architecture for cross-patient seizure type classification.
We used the recently released TUH EEG Seizure Corpus (V1.4.0 and V1.5.2) to evaluate the performance of SeizureNet. Experiments show that SeizureNet can reach a weighted $F1$ score of up to 0.94 for seizure-wise cross validation and 0.59 for patient-wise cross validation for scalp EEG based multi-class seizure type classification.
We also show that the high-level feature embeddings learnt by SeizureNet considerably improve the accuracy of smaller networks through knowledge distillation for applications with low-memory constraints. 
\end{abstract}
\section{Introduction}
Epilepsy is a neurological disorder which affects 1\% of the world's population. It causes sudden and unforeseen seizures which can result in critical injury, or even death of the patient. One third of epileptic patients do not have appropriate medical treatments available. For the remaining two thirds of the patients, treatment options and quality vary because seizure semiology is different for every epileptic patient. 
An important technique to diagnose epilepsy is through visual inspection of electroencephalography (EEG) recordings by physicians to analyse abnormalities in brain activities. This task is time-consuming and subject to inter-observer variability.
With the advancements in IoT-based data collection, machine learning based systems have been developed to capture abnormal patterns in the EEG data during seizures \cite{langkvist2014review,thodoroff2016learning,golmohammadi2017gated}. 
In this context, current systems have mostly focused on tasks such as seizure detection and seizure prediction \cite{vidyaratne2016deep,antoniades2016deep,lin2016classification}, and the task of seizure type classification is largely undeveloped due to factors such as complex nature of the task and unavailability of clinical datasets with seizure type annotations. Nevertheless, the capability to discriminate between different seizure types (e.g., focal or generalized seizures) as they are detected has the potential to improve long-term patient care, enabling timely drug adjustments and remote monitoring in clinical trials \cite{harrer2019artificial}. 
Recently, Temple University released TUH EEG Seizure Corpus (TUH-EEGSC) \cite{shah2018temple} for epilepsy research making it the world's largest publicly available dataset for seizure type classification.
The work of \cite{roy2019machine} presented baseline results on TUH-EEGSC \cite{shah2018temple} for seizure type classification by conducting a search space exploration of various standard machine learning algorithms. 
Other methods such as \cite{sriraam2019convolutional,saputro2019seizure} used subsamples of data from selected seizure types for seizure analysis.
%
%
In this paper, we propose an ensemble learning approach and present new benchmarks for seizure type classification using seizure-wise and patient-wise cross-validation.
The main contributions of this paper are as follows:
\begin{enumerate}
\item
We present \textit{SeizureNet}, a deep learning framework focused on diversifying individual classifiers of an ensemble by learning feature embeddings at different spatial and frequency resolutions of EEG data. Experiments show that our multi-spectral feature learning encourages diversity in the ensemble and reduces variance in the final predictions for seizure type classification.
\item
We present \textit{Saliency-encoded Spectrograms}, a visual representation which captures salient information contained in the frequency transform of the time-series EEG data. Experiments show that our saliency-encoded spectrograms produce improvements in seizure classification accuracy on TUH-EEGSC \cite{shah2018temple}.
\item
We evaluate the capability of our framework for transfering knowledge to smaller networks through knowledge distillation and present benchmark results for seizure type classification on TUH-EEGSC \cite{shah2018temple}. 
\end{enumerate}
\section{The Proposed Framework (SeizureNet)}\label{proposed_framework}
\begin{figure*}[t!]
  \begin{center}
    \includegraphics[trim=0.3cm 0.2cm 0.3cm 0.0cm,clip,width=1.0\linewidth,keepaspectratio]{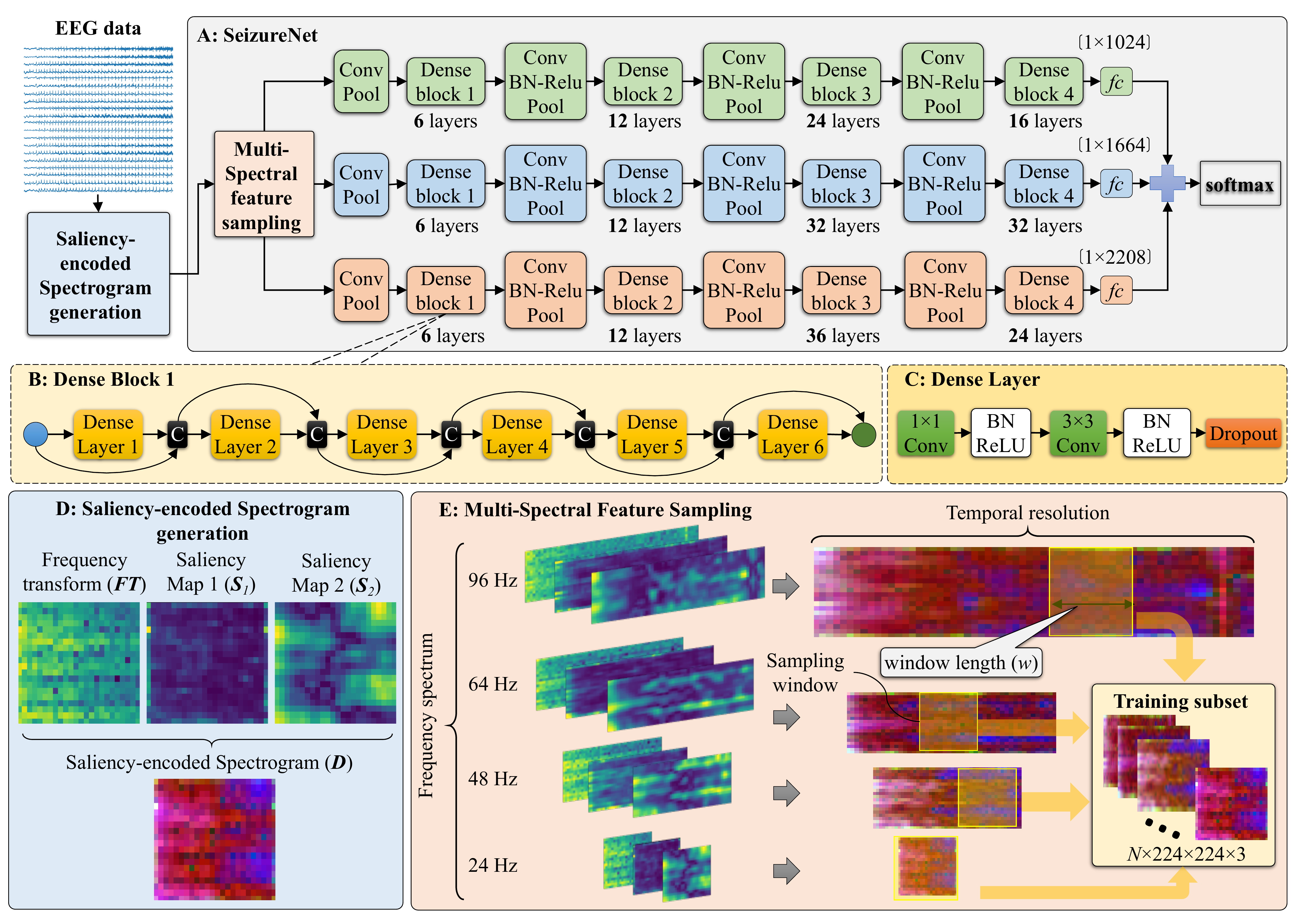}
  \vspace{-17pt}
  \caption{The overall architecture of our framework. Input EEG data is first transformed into the proposed saliency-encoded spectrograms (D), which are then sampled at different frequency and spatial resolutions (E), and finally fed into an ensemble of deep CNN models. For illustrative purposes, we show the ensemble with only three sub-networks. The outputs of the sub-networks are combined through summation and fed into a Softmax operation for producing probabilistic distributions with respect to the target classes (A).}
  \vspace{-20pt}
  \label{fig_network}
  \end{center}
\end{figure*}
Fig. \ref{fig_network}-A shows the overall architecture of our framework which transforms raw time-series EEG signals into the proposed saliency-encoded spectrograms, and uses an ensemble of deep CNN models to produce predictions for seizure type classification. 
In the following, we describe in detail the individual components of our framework.
\subsection{Saliency-encoded Spectrograms:}
Our saliency-encoded spectrograms are inspired from visual saliency detection \cite{hou2007saliency}, where we transform time-series EEG signals into a visual representation which captures multi-scale saliency information from the data.
Specifically, saliency-encoded spectrograms consist of three feature maps as shown in Fig. \ref{fig_network}-D. \textbf{i}) a Fourier Transform map ($FT$) which encodes the $log$ amplitude Fourier Transform of the EEG signals, \textbf{ii}) a spectral saliency map ($S_{1}$), which extracts saliency by computing the spectral residual of the $FT$ feature map, and \textbf{iii}) a multi-scale saliency map ($S_{2}$), which captures spectral saliency at multiple scales using center-surround differences of the features of the $FT$ feature map  \cite{montabone2010human,itti1998model}. 
Mathematically, given a time-series EEG sequence $\boldsymbol{X}(c,t)$ from a channel $c$ parameterized by time $t$, we compute the Fast Fourier Transform ($\mathcal{F}$) of the sequence as:
%
$\mathcal{F}(\boldsymbol{X})=\int_{\infty}^{-\infty}\boldsymbol{X}(c,t)e^{-2\pi it}dt.$
%
We compute $\mathcal{F}$ on data from selected 20 channels\footnote{The sclap EEG data was collected using 10-20 system \cite{silverman1963rationale}, and TCP montage \cite{lopez2016analysis} was used to select 20 channels of the input. 
We used the following 20 channels: $FP1-F7;F7-T3;T3-T5;T5-O1;FP2-F8;F8-T4;T4-T6;T6-O2;T3-C3;C3-CZ;CZ-C4;C4-T4;FP1-F3;F3-C3;C3-P3;P3-O1;FP2-F4;F4-C4;C4-P4;P4-O2$.
} and take the $log$ of the amplitude of the Fourier Transform. The output is reshaped into a $\mathbb{R}^{p\times 20}-$dimensional feature map ($FT$) where $p$ denotes the number of data points of the EEG sequence. Mathematically, $FT$ can be written as:
%
$FT=log(Amplitude(\mathcal{F}(\boldsymbol{X}))).$
%
Mathematically, $S_{1}$ can be written as:
%
$S_{1}=\mathcal{G}*\mathcal{F}^{-1}(\exp(FT-\mathcal{H}*FT)+\mathcal{P})^{2},$
%
where, $\mathcal{F}^{-1}$ denotes the Inverse Fourier Transform. The term $\mathcal{H}$ represents the average spectrum of $FT$ approximated by convoluting the feature map $FT$ by a $3\times3$ local averaging filter. The term $\mathcal{G}$ is a Gaussian kernel to smooth the feature values. The term $\mathcal{P}$ denotes the phase spectrum of the feature map $FT$.
The saliency map $S_{2}$ captures saliency in the feature map $FT$ with respect to its surrounding data points by computing center-surround differences at multiple scales. Let  $FT_{i}$ represents a feature value at location $i$, and $\Omega$ denotes a circular neighborhood of scale $\rho$ surrounding the location $i$. Mathematically, the saliency calculation at location $i$ can be written as:
%
$S_{2}(i)=\sum_{\rho\in[2,3,4]}(FT_{i}-min([FT_{k,\rho}])),\forall k\in\Omega,$
%
where, $[FT_{k,\rho}]$ represents the feature values in the local neighborhood $\Omega$.
Finally, we concatenate the three feature maps $FT$, $S_1$, and $S_2$ into an RGB-like data structure ($\mathcal{D}$) which is normalized between 0 and 255 range as shown in Fig. \ref{fig_network}-D. 
\subsection{Multi-Spectral Feature Learning:}\label{dense_sampling}
Deep neural networks are often over-parameterized and require sufficient amount of training data to effectively learn features that can generalize to the test data. When confronted with limited training data which is a common issue in health informatics \cite{alotaiby2014eeg}, deep architectures suffer poor convergence or over-fitting. 
To overcome these challenges, we present \textit{Multi-Spectral Feature Sampling} (MSFS), a novel method to encourage diversity in ensemble learning by training the sub-networks using data sampled from different frequency and temporal resolutions. 
Fig. \ref{fig_network}-E shows an overview of our MSFS method. 
Consider an $M-$dimensional training dataset $\boldsymbol{D}=\{(\mathcal{D}_i,y_i)|0\leq i \leq N_{d}\}$, which is composed of $N_{d}$ samples, where $\mathcal{D}_i$ is a training sample with the corresponding class label $y_i\in \mathcal{Y}$. During training, MSFS generates a feature subspace $\boldsymbol{D}^{m}=\{(\mathcal{D}_{i}^{m},y_i)|0\leq i \leq N_d\}$ which contains spectrograms generated by a random selection of the sampling frequency $f \in {F}$ (Hz), a window length parameter $w\in \mathcal{W}$ (seconds), and a window step size parameter $o\in\mathcal{O}$\footnote[2]{In this work, we used ${F}=[24,48,64,96]$ Hz, $\mathcal{W}=[1]$ second, and $\mathcal{O}=[0.5, 1.0]$ seconds.}. This process is repeated $N_{e}=3$ times to obtain the combination of random subspace $\{\boldsymbol{D}_{1}^{m},...,\boldsymbol{D}_{N_e}^{m}\}$, where $N_e$ is the size of the ensemble.
\subsection{The Proposed Ensemble Architecture (SeizureNet) :}\label{deep_network}
SeizureNet consists of $N_e$ deep Convolutional Neural Networks (DCNs). Fig. \ref{fig_network}-A shows the architecture of SeizureNet with three sub-networks. 
The basic building block of a DCN is a \textit{Dense Block} which is composed of multiple bottleneck convolutions interconnected through dense connections \cite{huang2016densely}. 
Specifically, each DCN model starts with a $7\times7$ convolution followed by Batch Normalization (BN), a Rectified Linear Unit (ReLU), and a $3\times3$ average pooling operation. Next, there are four dense blocks, where each dense block consists of $N_{l}$ number of layers termed \textit{Dense Layers} which share information from all the preceding layers connected to the current layer through fuse connections. Fig. \ref{fig_network}-B shows the structure of a dense block with $N_{l}=6$ dense layers. Each dense layer consists of $1\times1$ and $3\times3$ convolutions followed by BN, a ReLU, and a dropout block as shown in Fig. \ref{fig_network}-C.
Mathematically, the output of the $l^{th}$ dense layer in a dense block can be written as: 
%
$\mathcal{X}_l=[\mathcal{X}_0,...,\mathcal{X}_{l-1}],$
%
where $[\cdot\cdot\cdot]$ represents concatenation of the features produced by the layers $0,...,l-1$.
The final dense block produces $Y_{dense}\in \mathbb{R}^{k\times R\times7\times7}-$dimensional features which are squeezed to $k\times R-$dimensions through an averaging operation, and then fed to a linear layer $fc\in\mathbb{R}^{K}$ which learns probabilistic distributions of the input data with respect to $K$ target classes. 
To increase diversity among sub-networks of the ensemble, we vary the numbers of dense layers of Dense block 3 and Dense block 4 of the sub-networks.  
\subsection{Training and Implementation:}\label{implementation}
Consider a training dataset of spectrograms and labels $(D,y)\in (\boldsymbol{D}, \mathcal{Y})$, where each sample belongs to one of the $K$ classes $(\mathcal{Y}={1,2,...,K})$. 
The goal is to determine a function $f_{s}(D): \boldsymbol{D}\rightarrow \mathcal{Y}$. To learn this mapping, we train SeizureNet parameterized by $f(D,\theta^*)$, where $\theta^*$ are the learned parameters obtained by minimizing a training objective function:
%
$\theta^{*}=\argmin_{\theta}\mathcal{L}_{CE}(y,f(D,\theta)),$
%
where $\mathcal{L}_{CE}$ denotes a Cross-Entropy loss which is applied to the outputs of the ensemble with respect to the ground truth labels.
Mathematically, $\mathcal{L}_{CE}$ can be written as:
%
$\mathcal{L}_{CE}=\sum_{k=1}^K\mathbb{I}(k=y_i)\log\sigma (O_{e},y_i),$
%
where $O_e={1}/{N_e}\sum_{e=1}^{N_e}O_{k}$ denotes the combined logits produced by the ensemble, $O_k$ denotes the logits produced by an individual sub-network, $\mathbb{I}$ is the indicator function, and $\sigma$ is the SoftMax operation. It is given by: 
%
$\sigma(z_i)={\exp z_i}/{\sum_{k=1}^K \exp z_k}.$
%
For training the networks, we initialized the weights of the networks from zero-mean Gaussian distributions. The standard deviations were set to 0.01, and biases were set to 0. We trained the networks for 400 epochs with a start learning rate of 0.001 (which was divided by 10 at 50\% and 75\% of the total number of epochs), and a parameter decay of 0.0005 (on the weights and biases). 
Our implementation is based on the auto-gradient computation framework of the Torch library \cite{paszke2017automatic}. Training was performed by ADAM optimizer with a batch size of 50.
\section{Experiments and Results}\label{experiments}
\begin{table}[t!]
	\centering
	\caption{Statistics of TUH EEG Seizure Corp (v1.4.0 and v1.5.2) in terms of seizure types, number of patients and seizures count.}
	\label{table_tuh}
	\centering
	\setlength\tabcolsep{5.0pt}
	\begin{tabular}{@{}llcccc@{}}
		\toprule
		\multirow{2}{*}{Seizure type} &\multicolumn{2}{c}{Dataset v1.4.0}  &\multicolumn{2}{c}{Dataset v1.5.2} \\					
		&Patients  &{Seizures}&Patients  &{Seizures} \\					
		\midrule
		Focal Non-Specific (FN) &109&992&150&1836\\
		Generalized Non-Specific (GN) &44&{415}&81&583\\
		Simple Partial Seizure (SP)&2&44&3&52\\
		Complex Partial Seizure (CP)&34&{342}&41&367\\
		Absence Seizure (AB) &13&{99}&12&99\\
		Tonic Seizure (TN)&2& 67&3&62\\
		Tonic Clonic Seizure (TC)&11& 50&14&48\\
		Myoclonic Seizure (MC) & 3&2&3&2\\
		\bottomrule
	\end{tabular}
	\vspace{-10pt}
\end{table}
We used TUH EEG Seizure Corpus (TUH-EEGSC) \cite{shah2018temple} which is the  world’s  largest  publicly  available  dataset of  seizure  recordings with type annotations. TUH-EEGSC v1.4.0 released in October 2018 contains 2012 whereas TUH-EEGSC v1.5.2 released in May 2020 contains 3050 seizures. Table \ref{table_tuh} shows the statistics of TUH-EEGSC in terms of different seizure types and the number of patients.
For experiments, we excluded Myoclonic (MC) seizures from our study because the number of seizures was too low for statistically meaningful analysis (only three  seizures) as shown in Table \ref{table_tuh}.
For evaluations, we conducted cross validation at patient-level and seizure-level.
Specifically, for TUH-EEGSC v1.4.0, we considered seizure-wise cross validation. Since, TN and SP seizure types in TUH-EEGSC v1.4.0 contain data from only 2 patients, patient-wise cross validation will not yield statistically meaningful results. Hence, we considered 5-fold seizure-wise cross validation, in which the seizures from different seizure types were equally and randomly allocated to 5 folds.
For TUH-EEGSC v1.5.2, we considered patient-wise cross validation. Table \ref{table_tuh} shows that the 7 selected seizure types in TUH-EEGSC v1.5.2 comprise data from at least 3 patients, allowing statistically meaningful 3-fold patient-wise cross validation. In this scenario, the data was split into train and test subsets, so that the seizures in train and test subsets were always from different patients. This approach makes it more challenging to improve model performance but has higher clinical relevance as it supports model generalisation across patients.
\begin{table}[t!]
	\parbox{.48\linewidth}{
		\centering
		\caption{Average weighted f1 of SeizureNet and other methods on TUH-EEGSC v1.5.2 for patient-wise seizure type classification.}
		\label{table_classification}
		\vspace{-0pt}
		\centering
		\setlength\tabcolsep{15.0pt}
		\begin{tabular}{@{}lc@{}}
			\toprule
			\multirow{1}{*}{Methods} &\multicolumn{1}{c}{Weighted F1}\\	
			\midrule
			\cite{roy2019machine} Adaboost&0.473\\
			\cite{roy2019machine} SGD&0.432\\
			\cite{roy2019machine} XGBoost &0.561\\
			\cite{roy2019machine} kNN& 0.467\\
			\hline
			\rule{-2.5pt}{3ex}		
			SeizureNet (this work) &\textbf{0.592}\\
			\bottomrule
			\label{table_152}
		\end{tabular}
		\vspace{-12pt}
			\caption{Ablation study in terms of the proposed Multi-Spectral Feature Sampling (MSFS) using different sizes of training data.}
		\label{table_ablation_limited}
		\centering
		\setlength\tabcolsep{10.0pt}
		\begin{tabular}{@{}ccccccccccccc@{}}
			\toprule
			\multirow{1}{*}{Data} 
			&\multicolumn{2}{c}{Weighted F1 scores (seizure-wise)} \\
			size&without MSFS& with MSFS\\					
			\midrule
			10\%	&0.401		&\textbf{0.419}		\\
			20\%	&0.488		&\textbf{0.556}	\\
			50\%	&0.621		&\textbf{0.711}	\\
			\bottomrule
		\end{tabular}
	\vspace{-15pt}
	}
	\hfill
	\parbox{.47\linewidth}{
		\centering
		\caption{Average weighted f1 scores of SeizureNet and other methods on TUH-EEGSC v1.4.0 for seizure type classification at seizure-level.}
		\label{table_classification}
		\vspace{10pt}
		\centering
		\setlength\tabcolsep{1.0pt}
		\begin{tabular}{@{}lc@{}}
			\toprule
			\multirow{1}{*}{Methods} &\multicolumn{1}{c}{Weighted F1}\\	
			\midrule
			\cite{roy2019machine} Adaboost&0.593\\
			SAE (based on \cite{lin2016classification})&0.675\\
			LSTM (based on \cite{tsiouris2018long})&0.701\\
			CNN (based on \cite{acharya2018deep})&0.716\\
			\cite{roy2019machine} SGD&0.724\\
			CNN-LSTM (using \cite{thodoroff2016learning})&0.795\\
			CNN (based on \cite{sriraam2019convolutional})&0.802\\
			CNN (based on   \cite{o2018investigating})&0.826\\
			CNN-LSTM (using \cite{golmohammadi2017gated})&0.831\\
			\cite{roy2019machine} XGBoost &0.851\\
			\cite{roy2019machine} kNN& 0.898\\
			CNN (based on \cite{hao2018deepied})&0.901\\
			\hline
			\rule{-2.5pt}{3ex}
			SeizureNet (this work) &{0.940}\\
			\bottomrule
		\end{tabular}
		\vspace{-15pt}
	}
\end{table}
\begin{figure*}[t!]
	\begin{center}
		\includegraphics[trim=0.0cm 0.1cm 0.0cm 0.1cm,clip,width=0.9\linewidth,keepaspectratio]{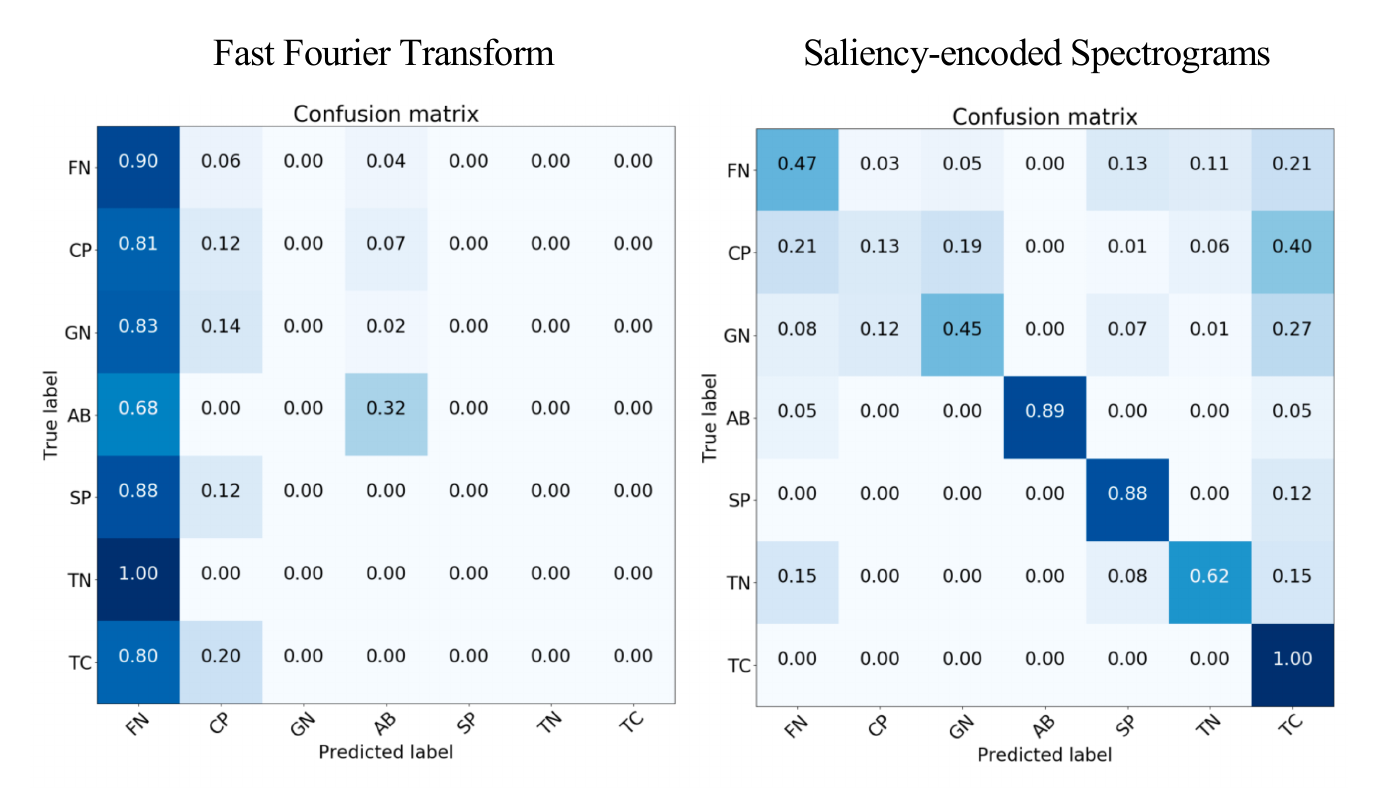}
		\vspace{-5pt}
		\caption{Confusion matrices produced by SeizureNet using Fourier transform and using the proposed saliency-encoded spectrograms, for 10\% of training data. The comparison shows that the confusions considerably decrease with the use of the proposed saliency-encoded spectrograms.}
		\vspace{-20pt}
		\label{fig_divspec}
	\end{center}
\end{figure*}
\begin{table}[t!]
		\caption{Ablation study of SeizureNet in terms of architecture, model parameters (million), FLOPS (million), inference speed, and Knowledge Distillation (KD).}
		\vspace{5pt}
		\centering
		\setlength\tabcolsep{4.0pt}\centering
		\begin{tabular}{@{}lcccccccc@{}}
			\toprule
			\multirow{1}{*}{Architecture} & {F1}&Param. &Flops &Time\\				
			\midrule
			SeizureNet&{0.940}&45.94&14241&90 ms\\
			ResNet&{0.727}&0.08&12.75&2 ms\\
			ResNet-KD&\textbf{0.746}&0.08&12.75&2 ms\\
			\bottomrule
		\vspace{-15pt}
		\label{table_ablation_arch}
		\end{tabular}
\end{table}
Table \ref{table_152} and Table \ref{table_classification} show patient-wise and seizure-wise cross validation results on TUH-EEGSC, respectively. The results show that SeizureNet improved patient-wise weighted f1 scores by around 3 points and seizure-wise weighted f1 scores by around 4 points compared to the existing methods. These improvements are mainly attributed to the proposed multi-spectral feature learning which captures information from different frequency and spatial resolutions, and enables SeizureNet to learn more discriminative features compared to the other methods. 
\subsection{SeizureNet for Knowledge Distillation:}
Here, we evaluated the capability of SeizureNet in transfering knowledge to smaller networks for seizure classification. For this, we trained a student ResNet model with 3 residual layers in conjunction with SeizureNet acting as a teacher network using a knowledge distillation based training function. 
Our training function $\mathcal{L}_{KD}$ is a weighted combination of a CrossEntropy loss term $\mathcal{L}_{CE}$ and a distillation loss term $\mathcal{L}_{KL}$. Mathematically, $\mathcal{L}_{KD}$ can be written as:
%
$\mathcal{L}_{KD}=\alpha\cdot \mathcal{L}_{CE}(P_{t},y)+\beta\cdot \mathcal{L}_{CE}(P_{s},y)+\gamma\cdot\mathcal{L}_{KL},$
%
where $P_{t}$ and $P_{s}$ represent the logits (the inputs to the SoftMax) of SeizureNet and the student model, respectively. The terms $\alpha \in [0,0.5,1]$, $\beta \in [0,0.5,1]$, and $\gamma \in [0,0.5,1]$ are the hyper-parameters which balance the individual loss terms. 
The distillation loss term $\mathcal{L}_{KL}$ is composed of Kullback-Leibler (KL) divergence function defined between log-probabilities computed from the outputs of SeizureNet and the student model.
Mathematically, $\mathcal{L}_{KL}$ can be written as:
$\mathcal{L}_{KL}(P_s,P_t)=\sigma(P_s)\cdot\left(\log(\sigma(P_s)) - \sigma(P_t)/T \right),$
%
where $\sigma$ represents the SoftMax operation and $T$ is a temperature hyper-parameter which controls the softening of the outputs SeizureNet. 
%
%
Table \ref{table_ablation_arch} shows that ResNet-KD produced improvement of around 2\% in the seizure-wise mean f1 score compared to the ResNet model without knowledge distillation. These results show that the feature embeddings learnt by SeizureNet can effectively be used to improve the accuracy of smaller networks (e.g., 3-layer ResNet having $45\times$ less training parameters, $1100\times$ less number of flops, and $45\times$ faster inference as shown in Table \ref{table_ablation_arch}), for deployment in memory-constrained systems.
\subsection{Significance of Saliency-encoded Spectrograms:}
\begin{figure*}[t!]
	\begin{center}
		\includegraphics[trim=0.1cm 0.0cm 0.1cm 0.1cm,clip,width=0.9\linewidth,keepaspectratio]{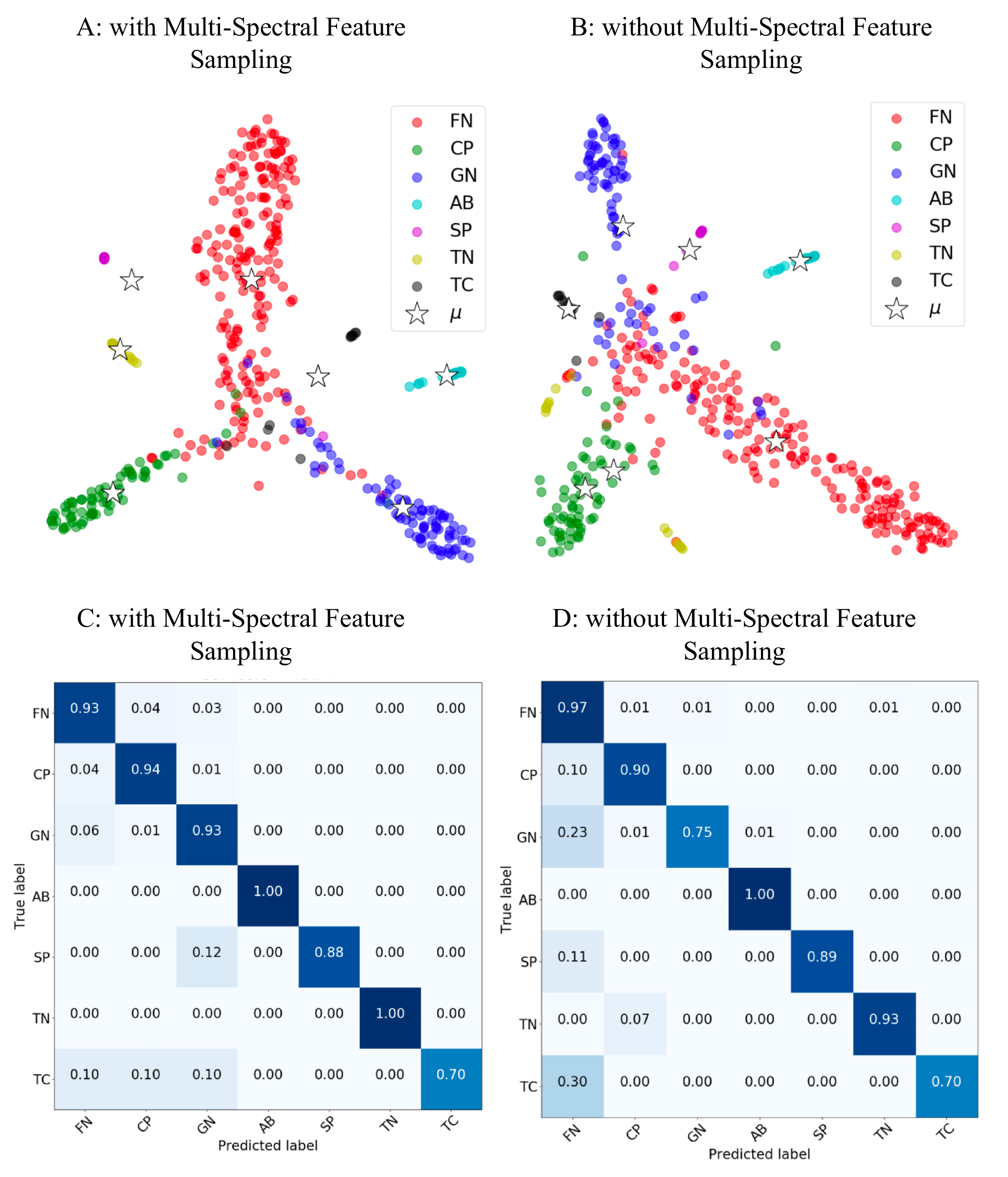}
		\vspace{-5pt}
		\caption{TSNE visualizations and confusion matrices of the seizure type manifolds produced by SeizureNet using the proposed multi-spectral feature learning (A, C) and without using the proposed multi-spectral feature learning (B, D), respectively. 
		}
		\vspace{-20pt}
		\label{fig_tsne}
	\end{center}
\end{figure*}
Fig. \ref{fig_divspec} shows that the combination of spectral residual of Fourier Transform and multi-scale center-surround difference information in the proposed saliency-encoded spectrograms turned out to be highly discriminative for seizure classification especially for small training data. 
For instance, when only 10\% of the data was used for training, the model trained using saliency-encoded spectrograms produced considerable decrease in the confusions for all the target classes compared to the model that was trained using only Fourier Transform. 
\subsection{Significance of Multi-Spectral Feature Learning:}
Table \ref{table_ablation_limited} shows that the models trained using MSFS produced higher f1 scores compared to the frequency-specific models. For instance, when only 50\% training data was used, models trained using MSFS improved seizure-wise f1 scores by around 9 points compared to the models trained without MSFS. These improvements show that information from different frequency bands compliment each other in better discriminating seizure classes especially for small data sizes compared to the features learnt at independent frequency bands. 
%
%
Fig. \ref{fig_tsne} shows a comparison of TSNE mappings produced by models trained with and without MSFS. The results show that the seizure manifolds produced with MSFS are better separated in the high-dimensional feature space (as shown in Fig. \ref{fig_tsne}-A) compared to the seizure manifolds produced without MSFS (as shown in Fig. \ref{fig_tsne}-B). This shows that increasing variation in the training information by combining data from different spatial and frequency bands is beneficial for learning discriminative features for seizure classification.
Fig. \ref{fig_tsne} also shows that models trained with MSFS produced less confusions (as shown in Fig. \ref{fig_tsne}-C), exhibiting the importance of combining data from different frequency and spatial resolutions.
\section{Conclusion and Future Work}
This paper presents a deep learning framework termed SeizureNet for EEG-based seizure type classification in patient-wise cross validation scenarios. 
The greatest challenge in a patient-wise validation approach is to learn robust features from limited training data which can effectively generalize to unseen test patient data. 
This is achieved through two novel contributions: i) saliency-encoded spectrograms which encode multi-scale saliency information contained in the frequency transform of the EEG signals, and ii) multi-spectral feature learning within an ensemble architecture, where spectrograms generated at different frequency and spatial resolutions encourage diversity in the information flow through the networks, and ensembling reduces variance in the final predictions. Experiments on the world's largest publicly available epilepsy dataset show that our SeizureNet produces competitive f1 scores for seizure type classification compared to the existing methods. 
Experiments also show that the feature embeddings learnt by SeizureNet considerably improve the accuracy of smaller networks through knowledge distillation. 
In future, we plan to investigate fusion of data from wearable sensors and videos for multi-modal seizure type classification in real-world epilepsy monitoring units. 
%
%
\bibliographystyle{splncs04}
\bibliography{ref}
\end{document}